\documentclass[review]{elsarticle}
\usepackage[utf8]{inputenc}

\usepackage{graphicx}
\usepackage{amsmath,amsfonts,amssymb}
\usepackage{mathtools}
\usepackage[version=4]{mhchem}
\usepackage{siunitx}
\usepackage{multirow}
\usepackage{longtable,tabularx}
\usepackage{bm}
\usepackage{caption}
\usepackage{subcaption}
\usepackage[export]{adjustbox}
\graphicspath{{Figures/}}
\usepackage[capitalise]{cleveref}
\DeclareMathOperator{\E}{\mathbb{E}}
\DeclareMathOperator{\Var}{Var}

\usepackage{tikz}

\begin{document}

\begin{frontmatter}
\title{Fusing CFD and measurement data using transfer learning}
\author[1]{Alexander Barklage}\ead{alexander.barklage@dlr.de}
\author[1]{Philipp Bekemeyer}\ead{philipp.bekemeyer@dlr.de}

\date{July 2025}

\affiliation[1]{organization={German Aerospace Center},
addressline={Lilienthalplatz 7},
postcode={38106},
city={Braunschweig},
country={Germany}}

\begin{abstract}
Aerodynamic analysis during aircraft design usually involves methods of varying accuracy and spatial resolution, which all have their advantages and disadvantages. It is therefore desirable to create data-driven models which effectively combine these advantages. Such data fusion methods for distributed quantities mainly rely on proper orthogonal decomposition as of now, which is a linear method. In this paper, we introduce a non-linear method based on neural networks combining simulation and measurement data via transfer learning. The network training accounts for the heterogeneity of the data, as simulation data usually features a high spatial resolution, while measurement data is sparse but more accurate. In a first step, the neural network is trained on simulation data to learn spatial features of the distributed quantities. The second step involves transfer learning on the measurement data to correct for systematic errors between simulation and measurement by only re-training a small subset of the entire neural network model. This approach is applied to a multilayer perceptron architecture and shows significant improvements over the established method based on proper orthogonal decomposition by producing more physical solutions near nonlinearities. In addition, the neural network provides solutions at arbitrary flow conditions, thus making the model useful for flight mechanical design, structural sizing, and certification. As the proposed training strategy is very general, it can also be applied to more complex neural network architectures in the future.
\end{abstract}
\end{frontmatter}

\section{Introduction}
Aircraft development requires high-fidelity aerodynamic data throughout the entire envelope which usually consists of computational fluid dynamics (CFD), wind tunnel, and flight test data. This data is important for many development aspects, such as flight mechanical design, structural sizing, and certification. However, these tasks require data for various flow conditions, so that the generation of high-fidelity data becomes infeasible. Hence, there is a desire to develop fast to evaluate surrogate models based on a reduced set of flow conditions. Ideally, these surrogate models combine information from all the aforementioned data sources to fuse the individual advantages into a single model. Corresponding methods have already been developed in the literature for scalar-valued quantities, commonly termed multi-fidelity methods, and for vector-valued quantities, termed data fusion methods in the following. Hereafter, the underlying models are called surrogates for scalar-valued quantities and reduced-order models for vector-valued quantities.

Multi-fidelity methods address the issue that high-fidelity data is usually more expensive to generate and therefore contains few data points. On the other hand, vast data is available for low-fidelity methods. The aim is to retain the accuracy of the high-fidelity model and derive trends from the low-fidelity models. Popular approaches are Kriging and Cokriging methods, as used by Han~\cite{Han.2012b,han2010new} for a RAE 2822 airfoil. Kriging methods can be divided into two classes, hierarchical and non-hierarchical models. Hierarchical models use a bridge function to incorporate trends from the low-fidelity models into the high-fidelity models~\cite{Han.2012}. These models can also be enhanced by taking gradient information into account~\cite{Han.2013}. Non-hierarchical Kriging models on the other side employ separate models for each fidelity level and combine these in a weighted sum~\cite{Feldstein.2020}. Kriging models have successfully been applied to various aerodynamic problems such as an inverted wing~\cite{Kuya.2011}, a low-boom supersonic jet~\cite{Choi.2008} and wing design~\cite{Forrester.2007,Keane.2003,Stradtner.2020}. Although these models provide excellent results for scalar-valued quantities, they are restricted to relatively low dimensions as the covariance matrices scale quadratically with the number of dimensions. Therefore, distributed quantities, such as the surface pressure distribution call for different models.

The curse of dimensionality of Kriging models for distributed quantities can be circumvented by reduced-order models. These types of models have been extensively investigated for single-fidelity data for decades. A popular model is a proper-orthogonal decomposition (POD) combined with an interpolation technique such as Gaussian process regression~\cite{Lucia.2004}. POD models have been successfully applied to aerodynamic applications to estimate loads \cite{Fossati.2015} and to speed up optimization problems \cite{Iuliano.2013, Ripepi.2018}. The POD has proven to provide accurate results for subsonic flows at a low training cost and straightforward use. However, once non-linear phenomena become dominant as is the case for transonic flows, the linear nature of the POD causes inaccurate predictions. Consequently, non-linear approaches are more suitable, and several examples exist in the literature that outperform POD models like Isomap~\cite{Franz.2014} and various deep-learning models~\cite{Yilmaz.2017, Sabater.2022, Hines.2023} while handling single-fidelity data.

Bertram et al.~\cite{Bertram.2018} applied reduced-order models to simulations of varying fidelity of a passenger car. They used a proper-orthogonal decomposition (POD) for dimension reduction and applied a Kriging model in the low-dimensional subspace. Renganathan et al.~\cite{Renganathan.2020} introduced a method that does not work with dimensionality reduction but uses a Bayesian approach where experimental and simulation data are combined in a weighted sum. However, both methods require that the dimensionality of all data is the same. This is generally not the case, as e.g. experimental pressure measurements usually have a coarser resolution than the CFD grid. Similar problems occur in image reconstruction, where a full-resolution image should be inferred from information of only a few pixels. An early approach for image reconstruction was introduced by Everson and Sirovich~\cite{Everson.1995}. They used a POD combined with a regression to estimate the POD basis coefficients, which best represent the gappy pixel information. Bui-Thanh et al.~\cite{BuiThanh.2004} adopted this method for aerodynamics to predict a pressure distribution from sparse data and termed it gappy POD. The gappy POD benefits from a regularization if a least-squares regression is used, as shown by Mifsud et al.~\cite{Mifsud.2019}. The same authors introduced a constrained version of the gappy POD where measured integral forces or moments can be incorporated as constraints. Bertram et al.~\cite{Bertram.2023} further improved this method by introducing Bayesian regression models, which can provide error estimates. A drawback of the gappy POD is that it produces nonphysical oscillations at nonlinearities, as the approach is inherently linear due to the POD. Hence, non-linear approaches such as artificial neural networks are gaining attention to overcome this issue. For example, shallow neural networks improve flow field predictions from sparse data~\cite{Erichson.2020}. Neural networks were successfully applied to infer flow field solutions from sparse data to turbulent flows~\cite{Guemes.2021} and blood flows~\cite{Arzani.2021}, among others. However, these approaches do not incorporate systematic discrepancies between the measurement and simulation data. To overcome this issue, transfer learning methods have already shown promising results in related problems such as inferring freestream parameters from pressure measurements~\cite{Yilmaz.2025} or predicting the pressure distribution for new flow conditions with limited data~\cite{Wang.2023}.

The novelty of this work is the application of transfer learning to data with different spatial discretizations and the demonstration of generalization for the whole spatial domain. We introduce a neural network approach that accounts for the heterogeneity of CFD and measurement data and efficiently combines the advantages of both data sources. Hence, the model retains the high spatial resolution of the CFD data while inheriting the high accuracy from the measurement data. It is also notable that the model is parametric, meaning it can provide a pressure distribution for conditions for which neither numerical nor experimental data are available, which is not possible with the established gappy POD approach. This is achieved by pre-training a neural network on CFD data followed by transfer learning on measurement data as will be described in section~\ref{sec:Methodology}. We demonstrate the method on data for the NASA common research model of different fidelity in section~\ref{sec:application}. The results are summarized in section~\ref{sec:conclusion} and an outlook is given.

\section{Neural network data fusion methodology}\label{sec:Methodology}

\begin{figure}[b!]
\centering
\includegraphics[width=.8\textwidth]{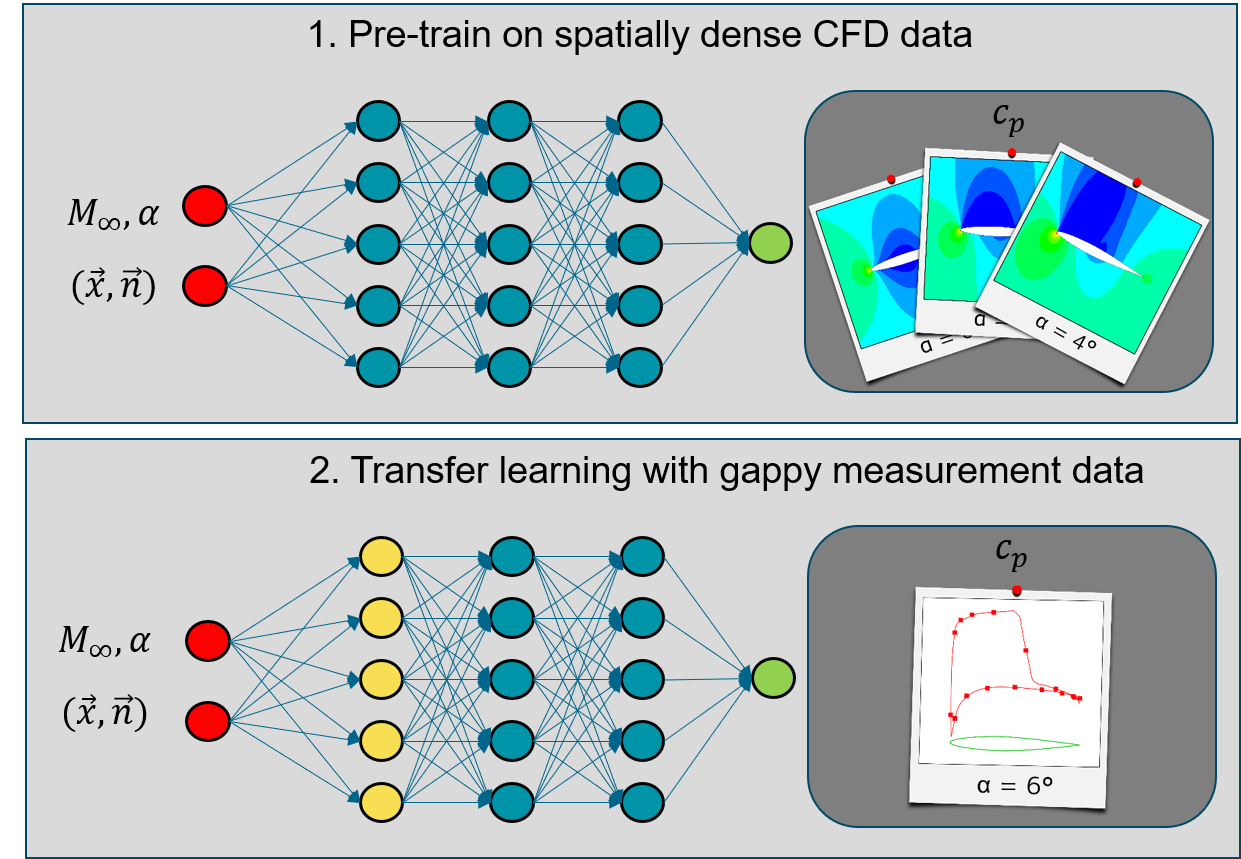}
\caption{Illustration of the neural network training. Trainable layers are colored blue and frozen layers are colored yellow.}
\label{fig:approach_sketch}
\end{figure}

The underlying idea of this work is to pre-train a neural network on CFD data and then perform transfer learning on sparse measurement data, as illustrated in Fig.~\ref{fig:approach_sketch}. In the pre-training phase, the model will learn spatial features of the flow from the dense CFD data, such as shocks or the stagnation line. Then, measurement data are used to adapt the model to overcome the modeling errors of the CFD. As the model should still remember the spatial information derived from the CFD data set, only parts of the network are re-trained with the measurement data. This approach is commonly known as transfer learning in the machine learning community. It is used when limited data is available for a specific task, but a large amount of data is available for a related task.

The neural network we are using here is an MLP with a constant dimension of the hidden layers, as illustrated in Fig.~\ref{fig:approach_sketch}. The network architecture is adopted from~\cite{Hines.2023}, where it showed comparable performance to more sophisticated graph neural networks (GNN). We chose this type of network over GNNs since the formulation is mesh-free and allows the evaluation of the model at arbitrary coordinates. Hence, it is not required that CFD data and measurements are available at the same locations. This is also a clear improvement over the gappy POD approach, which requires a nearest-neighbor interpolation. The input features of the MLP consist of aerodynamic parameters and geometric variables. We are using the freestream Mach number $M_\infty$ and the angle of attack $\alpha$ as aerodynamic input parameters, though an arbitrary parametrization can be chosen. The geometrical inputs are the spatial coordinates $\Vec{x}=(x,y,z)$ and the surface normals $\Vec{n}=(n_x,n_y,n_z)$ as only surface quantities are of interest here. A min-max scaler transforms the inputs to lie in the interval $[-0.5,0.5]$. The output of the neural network is the pressure coefficient at the input coordinates.

The training approach consists of two steps, as illustrated in Fig.~\ref{fig:approach_sketch}. In the first step, the MLP is trained on spatially dense CFD data for different flow conditions that ideally cover the parameter space of interest. Each grid point of the CFD mesh corresponds to a single training sample, resulting in a rich training data set. Consequently, the neural network learns the spatial structures of flow features at a high resolution. The network dimensions are determined via a hyperparameter optimization and will be kept fixed in the second training step. The hyperparameters were optimized using a Bayesian algorithm based on Gaussian processes with 36 randomly chosen initial points and 100 total trials. The optimization considers five hyper parameters which are the learning rate, the dimension of the hidden layers, the number of hidden layers, and the learning rate decay factor. All hidden layers apply the same activation function (elu) except for the last layer, which is linear. A learning rate scheduler decreases the learning rate following an exponential law according to the following expression
\begin{equation}
        l_r= 
\begin{cases}
    l_0 \cdot\gamma^{\text{epoch}-9},& \text{if } \text{epoch}\geq 10\\
    l_0,              & \text{otherwise}
\end{cases}
\end{equation}

After the network architecture is fixed and the neural network is trained on CFD data in the first training step, transfer learning is applied with measurement data. As highlighted in Fig.~\ref{fig:approach_sketch}, only part of the network will be retrained while the weights and biases of the first layers are kept frozen. This ensures that the network retains the spatial features it learned from the CFD data. Another hyperparameter optimization is performed to determine the number of retraining layers, the learning rate, and the learning rate decay factor.

\section{Gappy POD for data fusion}

This section will briefly describe the established gappy POD method for data fusion, as it serves as a performance comparison case in the following. For mathematical details, the reader is referred to~\cite{Bertram.2023}. The core of the method is a POD model generated from $n$ CFD solutions. These solutions are accumulated in the so-called snapshot matrix $Y\coloneqq [y^1,\dots,y^n] \in \mathbb{R}^{N\times n}$, where $N$ is the number of points in the grid. Without loss of generality, we assume that the data is normalized by subtracting the mean. Applying a singular value decomposition to the snapshot matrix gives
\begin{equation}
    Y = U\Sigma V^T~,
\end{equation}
where $U\in \mathbb{R}^{N\times N}$ and $V \in \mathbb{R}^{n\times n}$ are orthogonal matrices and $\Sigma \in \mathbb{R}^{N\times n}$ is a diagonal matrix containing the singular values. The singular values quantify the information content of the corresponding vector in the matrix $U$. It is therefore possible to approximate the snapshots using a reduced set of modes without losing relevant information content. The POD basis corresponds to the set of $r$ reduced modes $U_r\coloneqq \{u^1,\dots , u^r\}\in \mathbb{R}^{N\times r}$. The original snapshots can then be approximated by the POD basis as 

\begin{equation}
    y^i \approx U_r \hat{a}~,
\end{equation}
where $\hat{a} = (\hat{a}_1,\dots, \hat{a}_r)$ are the POD basis coefficients.

The assumption of gappy POD is that measurements can be approximated in the same way with suitable POD basis coefficients. Measurements are assumed to be sparse so that the measurement vector $Y\in \mathbb{R}^{m}$ with $m<N$ can be written as
\begin{equation}\label{eq:gappyPOD_linearRegression}
    y \approx L(\delta) U_r \hat{a} +\epsilon~,
\end{equation}
where $L(\delta) \in \mathbb{R}^{N\times m}$ is an observation matrix for given sensor positions $\delta$ and $\epsilon \sim \mathcal{N}(0,\Sigma)$ is measurement noise with covariance matrix $\Sigma \in \mathbb{R}^{m\times m}$. The observation matrix is a mask matrix consisting of standard basis vectors $e_{j_i}\in \mathbb{R}^{N}, i=1, \dots,m$ where $j_i$ is the grid index where measurements are available. The grid indices $j_i$ are determined by a nearest-neighbor search since the CFD grid is usually spatially dense.
Equation~\ref{eq:gappyPOD_linearRegression} provides a linear regression formulation for determining the POD coefficients. Here, we reformulate the problem in a form suitable for Gaussian process regression with $x=L(\delta) U_r$ as input. As detailed in~\cite{Bertram.2023}, the posterior predictive distribution then reads
\begin{align}
    \E [f_*] &= k(x_*)(K+\sigma^2)^{-1}Y \\
    \Var [f_*] &= k(x_*,x_*) - k(x_*)^T(K+\sigma^2)^{-1}k(x_*)
\end{align}
where $x_* \equiv (U_r)_i, i=1,\dots,N$ is the full POD basis,  $k(x_*)\coloneqq (k(x_*, x_i))_{i=1,\dots m}$, and $K\coloneqq (k(x_i, x_j))_{i,j=1,\dots m}$. Hence, the posterior predictive solution is defined at the CFD grid points in contrast to the neural network approach where the resolution can be arbitrary. The kernel function $k(x, x')$ we are using is a combination of a radial basis function kernel and dot product kernel
\begin{equation}
    k(x, x')=\theta_0\cdot \exp(-\theta_1 \lVert x-x' \rVert ^2) + \theta_2 x^Tx'
\end{equation}

The hyperparameters $\theta_0, \theta_1, \theta_2$ are optimized by maximizing the log marginal likelihood function.

\section{Application example}\label{sec:application}

We demonstrate the performance of the neural network data fusion approach on data for the common research model. Results are compared to the gappy POD method in terms of accuracy and computational time. The CFD training data set consists of rigid simulations, although the test case features significant aero-elastic effects. We purposely ignored these effects to introduce an epistemic error between the CFD and measurements. First, we apply the data fusion to synthetic measurement data extracted from simulations including aero-elastic effects. These results are presented in Sec.~\ref{sec:results_CFD_CSM}. Afterwards, the method is demonstrated on real measurement data in Sec.~\ref{sec:results_measurements}. Two different training strategies are applied for a fair comparison with the gappy POD method. The first is referred to as single-point strategy and only uses measurement data for one flow condition and predicts the surface solution at high spatial resolution for the same flow condition which is analogous to the gappy POD approach. The second strategy termed multi-point strategy uses measurements for multiple flow conditions for transfer learning, while predicting for an unseen flow condition. This is not possible with the gappy POD and is a clear advantage since predictions can be made for flow conditions for which no measurement data is available. Although it is possible to build another ROM on top of the gappy POD results to get similar characteristics, this will introduce additional model inaccuracies and is regarded beyond the scope of this paper.

\subsection{Test case description}\label{sec:application_test_case_description}

\begin{figure}[b!]
\centering
\begin{subfigure}[t]{0.48\textwidth}
    \centering
    \includegraphics[width=\textwidth]{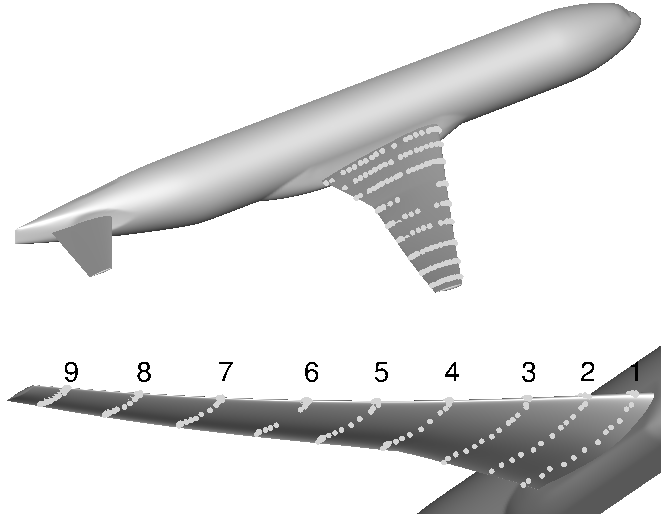}
    \caption{Overview on the sensor locations.}
    \label{fig:application_data_NASA_CRM}
\end{subfigure}
\hfill
\begin{subfigure}[t]{0.48\textwidth}
    \includegraphics[width=\textwidth]{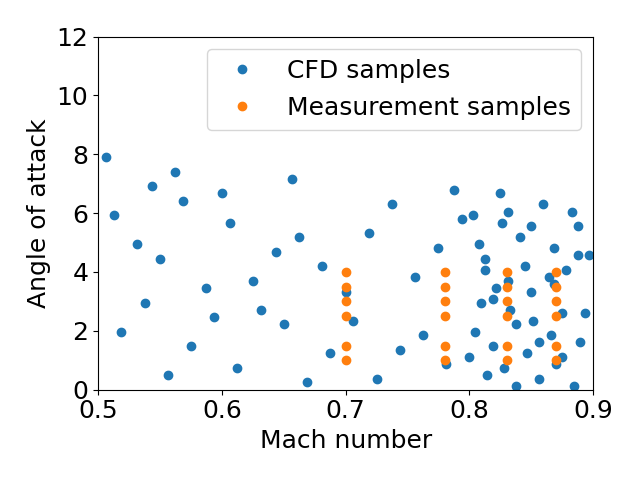}
    \caption{Flow conditions at which experimental and numerical data is available.}
    \label{fig:appl_NASA-CRM_DoE}
\end{subfigure}        
\caption{Overview on the available experimental and numerical data sets.}
\label{fig:application_data_set_overview}
\end{figure}
The NASA common research model is an extensively studied aerodynamic test case introduced by Vassberg et al.~\cite{Vassberg.2008} for which various configurations are available focusing on different aerodynamic aspects. For the present study, we chose the high-speed configuration with a vertical tail, which is depicted in Fig.~\ref{fig:application_data_NASA_CRM}. The gray points highlight the 253 pressure sensor positions which are distributed over 9 spanwise sections. The CFD training data covers a two-dimensional parameter space of $(M,\alpha)\in [0.5,0.9]\times[\SI{0}{\degree},\SI{10}{\degree}]$. A Halton sampling was applied to explore the parameter space, leading to the parameter combinations highlighted by blue dots in Fig.~\ref{fig:appl_NASA-CRM_DoE}. As transonic flow conditions are more challenging to model, the sample density is higher in the Mach number range of $M\in [0.8,0.9]$, resulting in a total number of available CFD samples of 80.

The simulations rely on a computational grid, originally used by Keye et al.~\cite{Keye.2017}, which contains $22\times10^6$ grid points. The numerical data comprises rigid simulations and coupled fluid-structure simulations for the 80 parameter combinations described above. The coupled fluid-structure simulations apply an iterative process of performing CFD simulations and a modal structural solver. In each step, aerodynamic loads and structural deformations are exchanged between the two disciplines. For a detailed description of the coupled fluid-structure simulations, the reader is referred to~\cite{Barklage2024}. All CFD simulations considered here are RANS computations, performed with the DLR TAU code~\cite{Kroll.2014}. Spatial discretization applies a central scheme with artificial matrix dissipation for convective terms and second-order central differences for viscous terms. Time integration is performed by an implicit backward Euler scheme. The turbulence model used is the one-equation Spalart-Allmaras turbulence model ~\cite{SPALART.1992}.

The experimental data are taken from measurements in the European Transonic Wind Tunnel (ETW) in 2014~\cite{Boyet.2018}. From these data, we selected 16 different flow conditions at four Mach numbers of $M_\infty = \{0.7, 0.78, 0.83, 0.87\}$ and four angles of attack $\alpha = \{1, 1.5, 2.5, 3.5\}\si{\degree}$, as illustrated in Fig.~\ref{fig:appl_NASA-CRM_DoE}. The Reynolds number is $Re=\SI{1e6}{}$ for all samples, corresponding to the CFD simulations. The total pressure and total temperature are also constant for all investigated cases at $T_0=\SI{300}{\kelvin}$ and $p_0=\SI{191}{\kilo\pascal}$.

\subsection{Hyperparameter optimization and training of the neural network}

The neural network is pre-trained on the rigid CFD data as described in section~\ref{sec:Methodology}. Each grid point for each flow condition represents one input sample, resulting in a data set consisting of approximately 10 million samples. This data set is divided into a training and validation set by a \SI{80}{\percent}/\SI{20}{\percent} random split to achieve a good generalization of the neural network in the spatial domain and for different flow conditions. The hyperparameter optimization took approximately 55 hours on a single GPU (NVIDIA RTX A4000) for a batch size of 4096. Table~\ref{tab:NN_hyperparameters_pretrain} lists the optimal hyperparameters that define the architecture of the neural network. Using these parameters, the network is trained on the complete data set including the validation samples.

The hyperparameter optimization of the fine-tuning step is based on the synthetic measurement data which is described in section~\ref{sec:results_CFD_CSM}. The training data set contains 8096 samples, which is three orders of magnitude smaller than for pretraining. Hence, it is essential to reduce the learning rate to prevent overfitting. As listed in Table~\ref{tab:NN_hyperparameters_pretrain}, the optimal learning rate of the fine-tuning step is almost two orders of magnitude smaller than that of the pre-training step. It is also necessary to fine-tune most parts of the network, as only the first two layers are frozen, leaving approximately \SI{84}{\percent} trainable parameters. The hyperparameters listed in Table~\ref{tab:NN_hyperparameters_pretrain} will be used for all the following fine-tuning tasks even if the training data is different since it is infeasible to always perform a new hyperparameter optimization in practical applications.

\begin{table}[h]
    \caption{Optimal hyperparameters for the two training stages.}
    \centering
    \begin{tabular}{lll}
        \hline
        Training stage & Hyperparameter & Value \\
        \hline
        \hline
        Pre-training & Initial learning rate & \SI{1e-3}{} \\
         & LR decay factor & 0.995 \\
         & Dimension of hidden layers & 64 \\
         & Number of hidden layers & 9 \\
         \hline
         Fine tuning & Initial learning rate & \SI{3e-5}{} \\
         & LR decay factor & 0.998 \\
         & Number of fine-tuning layers & 7 \\
    \end{tabular}
    \label{tab:NN_hyperparameters_pretrain}
\end{table}

\begin{figure}[t!]
\centering
\begin{subfigure}[t]{0.48\textwidth}
    \includegraphics[width=\textwidth]{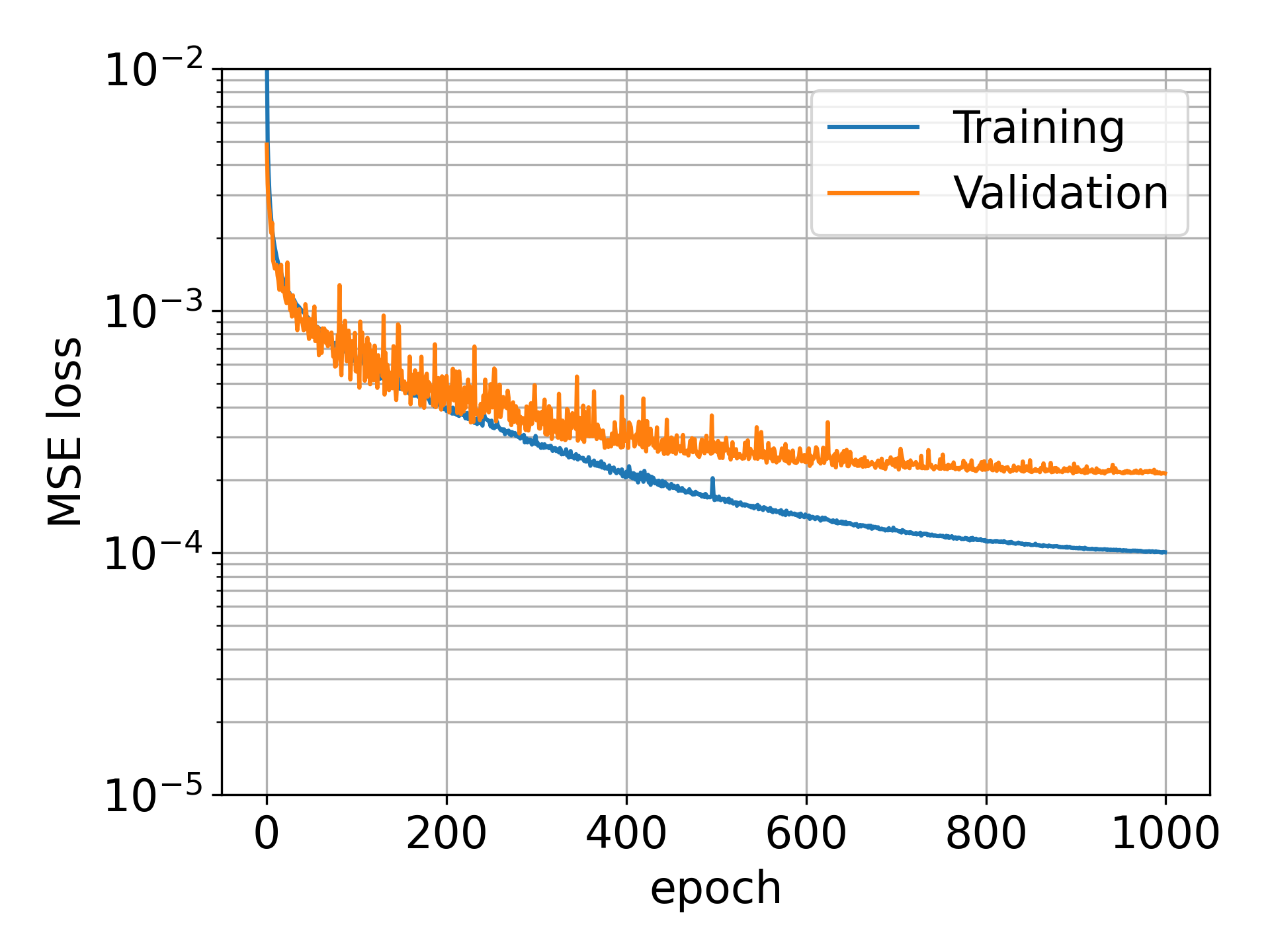}
    \caption{Loss for the pre-training step.}
    \label{fig:training_history_pretrain}
\end{subfigure}
\hfill
\begin{subfigure}[t]{0.48\textwidth}
    \centering
    \includegraphics[width=\textwidth]{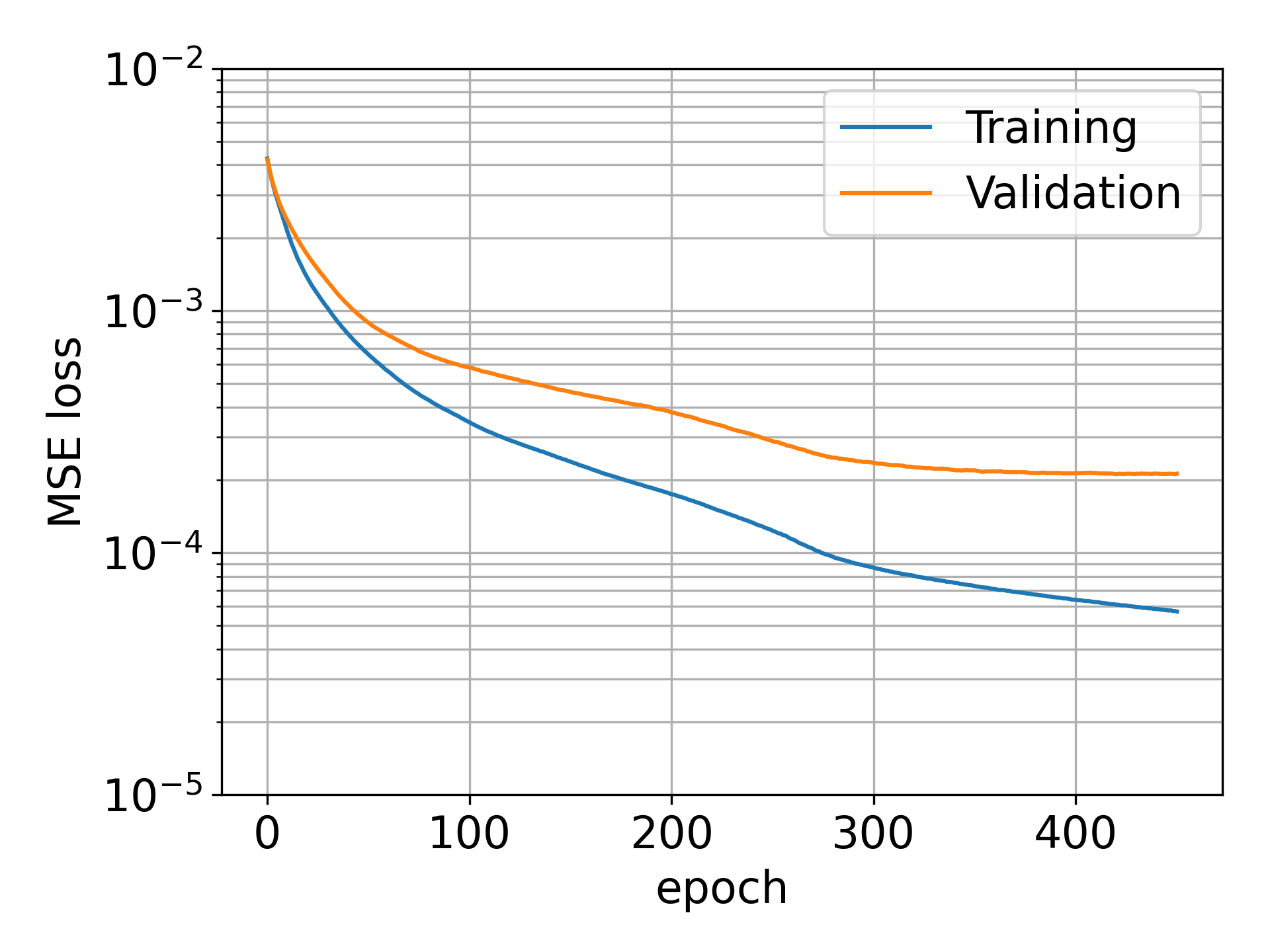}
    \caption{Loss for the fine-tuning step.}
    \label{fig:training_history_finetune}
\end{subfigure}
\caption{Evolution of the loss during the two training stages.}
\label{fig:training_history}
\end{figure}

Overfitting is prevented by using an early stopping algorithm with patience of 100 epochs for pre-training and 30 epochs for fine-tuning. The resulting training histories for pre-training and fine-tuning are shown in Fig.~\ref{fig:training_history}. The pre-training phase takes 1000 epochs to reach a converged validation loss. Up to approximately 600 epochs, the validation loss features an oscillatory behavior that diminishes at the end of training due to the learning rate decay. In contrast, the validation loss during fine-tuning shows a much smoother evolution as a consequence of the small learning rate. The validation loss reaches a plateau at the end of training just as in the pre-training step, showing that the model does not overfit.

\subsection{Validation on high-fidelity CFD data}\label{sec:results_CFD_CSM}

\begin{figure}[b!]
\centering
\includegraphics[width=.72\textwidth]{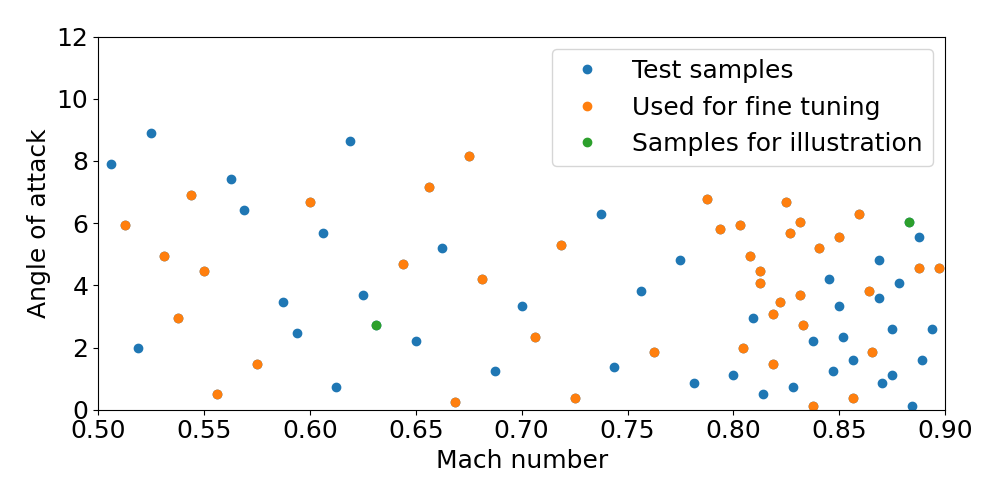}
\caption{Chosen sample points for fine tuning and testing.}
\label{fig:validation_CFD_sample_points}
\end{figure}

We will compare the performance of the data fusion methods based on synthetic measurements, extracted from the high-fidelity aero-elastic simulations. This allows to quantify the accuracy of the fused solution on the whole wing surface which is not possible solely relying on the sparse measurements. As described in Section~\ref{sec:application}, two training strategies are applied for the neural network. The strategy that is fine-tuned on multiple flow conditions uses the samples highlighted in orange in Fig.~\ref{fig:validation_CFD_sample_points}. The remaining blue sample points are used for testing, while the two green points are chosen to illustrate the results. The first of these two samples is an interpolation case with a maximum distance to the nearest fine-tuning samples at subsonic flow conditions of $M_\infty=0.63$ and $\alpha=\SI{2.7}{\degree}$. The second sample corresponds to a transonic case at $M_\infty=0.88$ and $\alpha=\SI{6}{\degree}$ that lies at the border of the parameter space, thus corresponding to an extrapolation case.

\begin{figure}[b!]
\centering
\includegraphics[width=\textwidth]{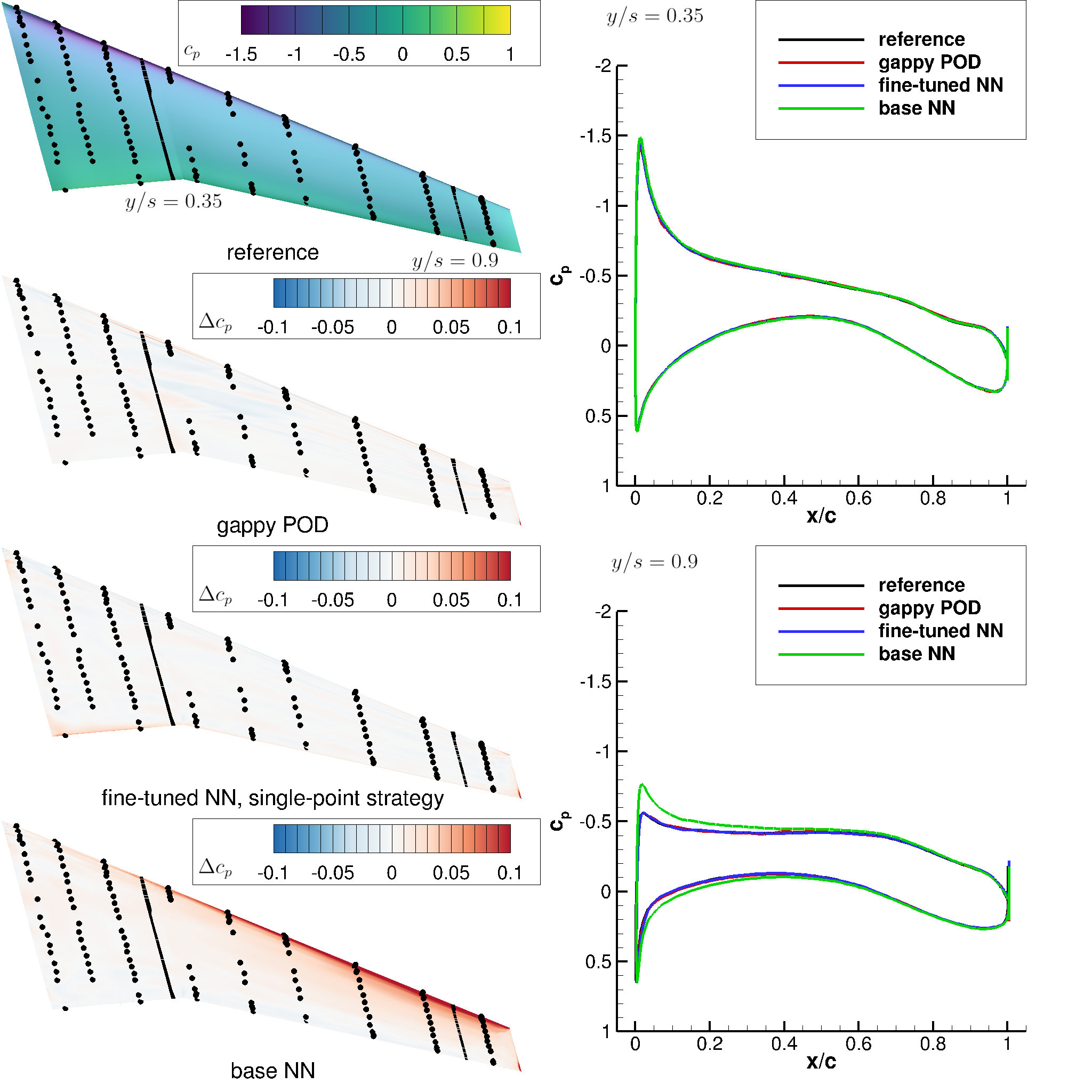}
\caption{Comparison of the approximation errors for $M_\infty=0.63$ and $\alpha=\SI{2.7}{\degree}$.}
\label{fig:validation_CFD_error_contour_sample34}
\end{figure}

\begin{figure}[b!]
\centering
\includegraphics[width=\textwidth]{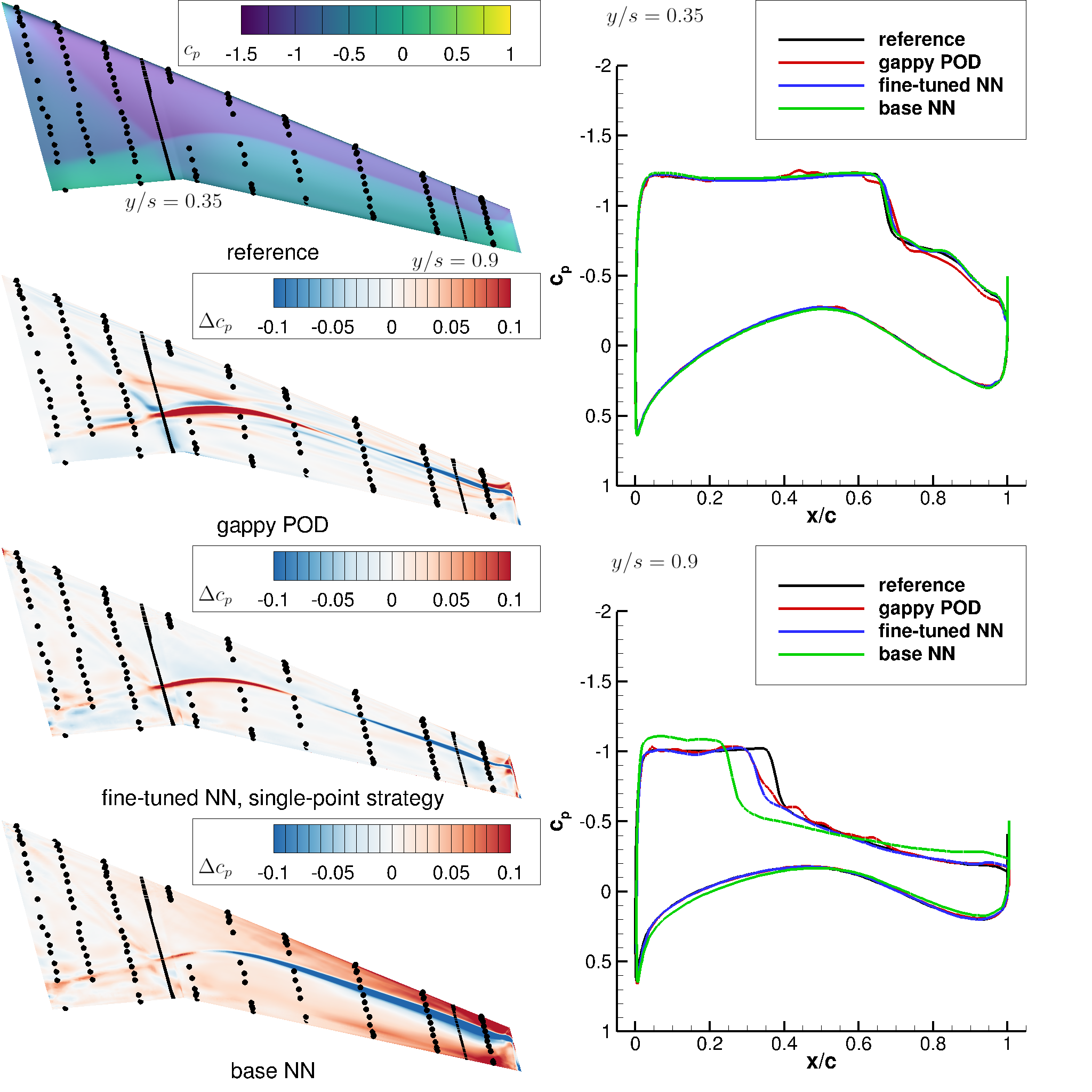}
\caption{Comparison of the approximation errors for $M_\infty=0.88$ and $\alpha=\SI{6}{\degree}$.}
\label{fig:validation_CFD_error_contour_sample79}
\end{figure}

Figure~\ref{fig:validation_CFD_error_contour_sample34} shows the results of the gappy POD, the fine-tuned neural network, and the base neural network for the subsonic case. The base neural network refers to that which is trained on the rigid data and is not fine-tuned. The first training strategy is chosen here for comparison, since it is analogous to the gappy POD approach. The left part of the figure depicts the reference solution and the deviations to it for the three different predictions. As is evident from the errors of the base neural network, deviations from the rigid training data are small at the wing root and increase in the spanwise direction. This is due to the bending and twisting of the wing, which increase in spanwise direction, causing a change in pressure distribution. Both data fusion models compensate for these systematic errors, resulting in a similar accuracy. A closer comparison is shown in the right part of the figure for two spanwise cuts near the kink and close to the wing tip. Both cuts lie between two measurement sections to illustrate the interpolation capabilities in space. At the innermost cut, all solutions coincide as deviations to the rigid solution are small. At the outermost cut, the reference solution features a less pronounced suction peak than the rigid solution due to a decreased local angle of attack caused by the deformations. The data fusion methods adapt to the reference solution based on what they infer from the measurements, resulting in negligible differences.

The transonic case is more challenging, exhibiting non-linear flow features such as shocks. As shown in the top left of Fig.~\ref{fig:validation_CFD_error_contour_sample79}, the pressure distribution is dominated by a complex shock system. It is essential to precisely capture the location and strength of the shock as it significantly influences structural loads and performance quantities such as lift and drag. Neglecting structural deformations leads to a profoundly different shock location, as the comparison to the base neural network shows. This effect becomes more pronounced in the spanwise direction. Whereas the shock location is almost the same at the wing root, it is shifted downstream at the wing tip. Both data fusion methods improve the prediction of the shock location at the tip. However, the prediction is worse at the kink, due to a lack of sensors in close proximity to the shock. The gappy POD features errors larger than the fine-tuned neural network, which are not confined to the shock region. As already observed in prior studies, the gappy POD solution exhibits oscillations upstream and downstream of the shock. This is illustrated in the section cuts on the right in Fig.~\ref{fig:validation_CFD_error_contour_sample79}. At $y/s=0.35$, the gappy POD produces nonphysical solutions, although the rigid training data already represents the reference data quite well. In contrast, the neural network closely follows the reference with a more physical solution. The same conclusions apply for the cut at $y/s=0.9$ where the neural network predicts a sharper shock. However, the shock location is predicted too far upstream, which is presumably related to a low sensor density in the shock region.

For a quantitative comparison, we use the area-weighted root mean squared error (RMSE) which is summarized in Teble~\ref{tab:validation_CFD_errorMetrics} for the two selected conditions and the whole test data set. For the subsonic test case, the gappy POD performs best, followed by the fine-tuned neural net following the single-point strategy. However, the multi-point strategy provides almost the same accuracy as single-point strategy although it has not seen the measurement data for this condition. Both training strategies lead to RMSE values, which are about four times smaller than the base neural network. For the transonic case, the error reduction is smaller but still significant. Again, both training strategies lead to similar errors, showing that the multi-point strategy generalizes well. As observed earlier, the gappy POD leads to higher errors than the neural network since it struggles handling non-linearities. By evaluating the whole test data set, the fine-tuned neural networks perform better than the gappy POD and lead to the same RMSE value. Hence, the neural network can be fine-tuned with a small measurement data set and still outperforms the gappy POD for unseen data.

\begin{table}[h]
    \centering
    \begin{tabular}{l|llll}
        \hline
        Data set & \multicolumn{4}{c}{RMSE} \\
         & gappy POD & SP & MP & base NN \\
        \hline
        \hline
        $M_\infty=0.63, ~\alpha=\SI{2.7}{\degree}$ & \SI{4.31e-3}{} & \SI{5.71e-3}{} & \SI{5.77e-3}{} & \SI{2.33e-2}{} \\
        $M_\infty=0.88,~\alpha=\SI{6.0}{\degree}$ & \SI{2.66e-2}{} & \SI{2.08e-2}{} & \SI{2.14e-2}{} & \SI{5.07e-2}{} \\
        Entire test data set & \SI{1.87e-2}{} & \SI{1.47e-2}{} & \SI{1.47e-2}{} & \SI{4.09e-2}{} \\
    \end{tabular}
    \caption{Root mean squared error of fused results. SP: single-point strategy, MP: multi-point strategy}
    \label{tab:validation_CFD_errorMetrics}
\end{table}

\subsection{Validation on measurement data}\label{sec:results_measurements}
In this section, we demonstrate the performance on real measurement data. However, in contrast to the previous section, only the measurement points can be used for validation. Therefore, we will only use a subset of the measurements for fine-tuning leaving out sections 3 and 9. The multi-point strategy uses the 16 measurement samples described in Section~\ref{sec:application_test_case_description}. The validation samples cover two Mach numbers of $M_\infty = \{0.7, 0.85\}$ and two angles of attack $\alpha = \{2.0, 4.0\}\si{\degree}$. Note that these parameter combinations are not contained in the training set for fine-tuning and that $\alpha=\SI{4}{\degree}$ is an extrapolation case.

\begin{figure}[b!]
\centering
\begin{subfigure}{0.49\textwidth}
    \includegraphics[width=\textwidth]{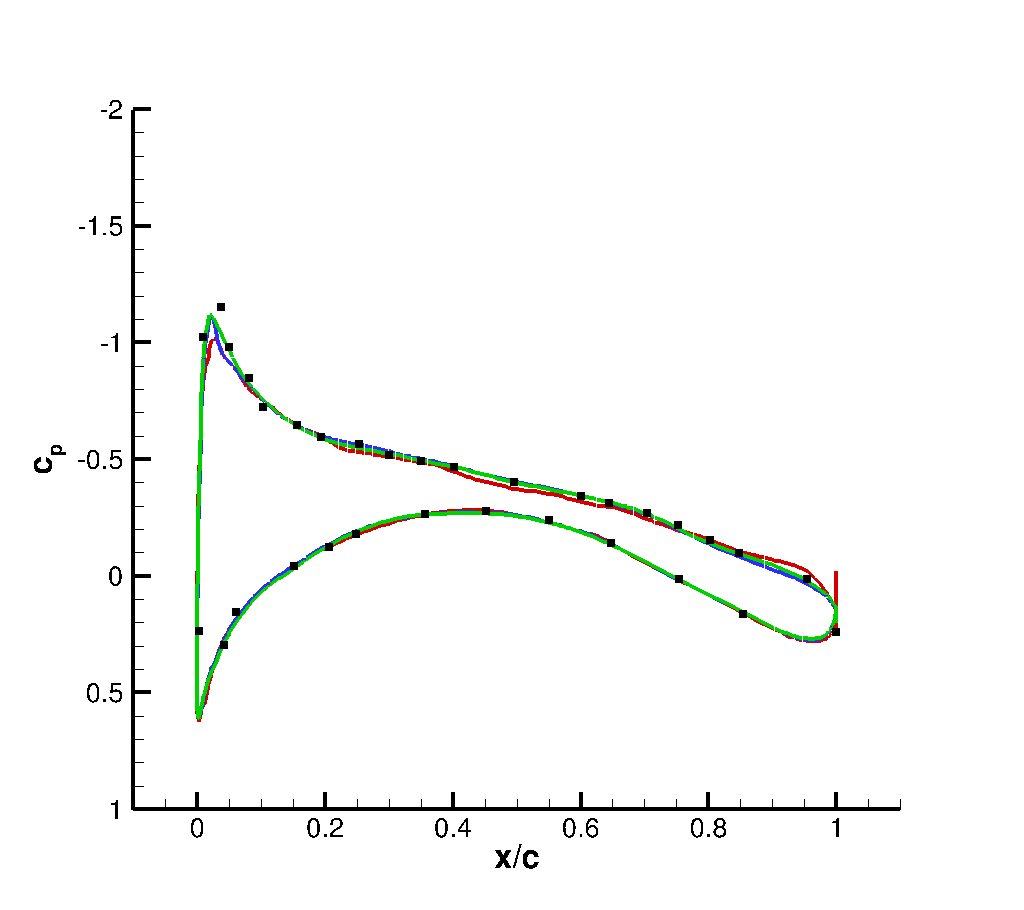}
    \caption{$\alpha = 2.0\si{\degree}$, section 3.}
    \label{fig:first}
\end{subfigure}
\hfill
\begin{subfigure}{0.49\textwidth}
    \includegraphics[width=\textwidth]{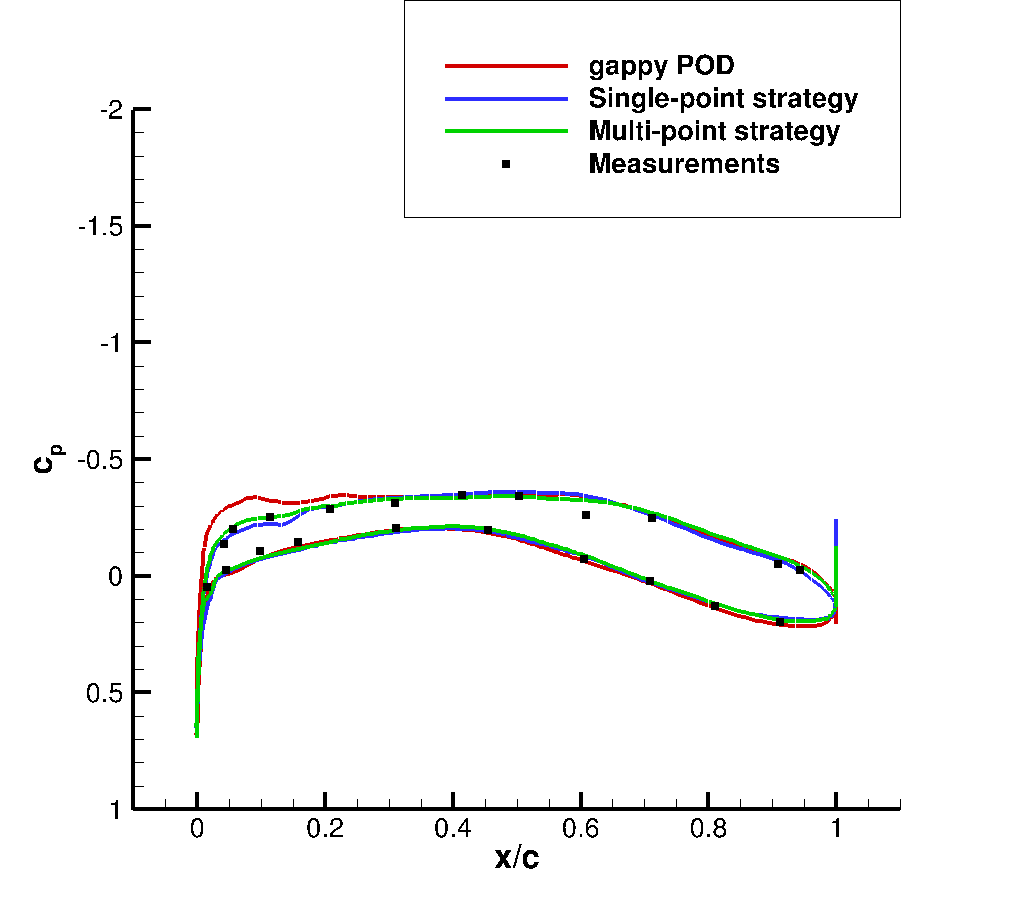}
    \caption{$\alpha = 2.0\si{\degree}$, section 9.}
    \label{fig:second}
\end{subfigure}
\begin{subfigure}{0.49\textwidth}
    \includegraphics[width=\textwidth]{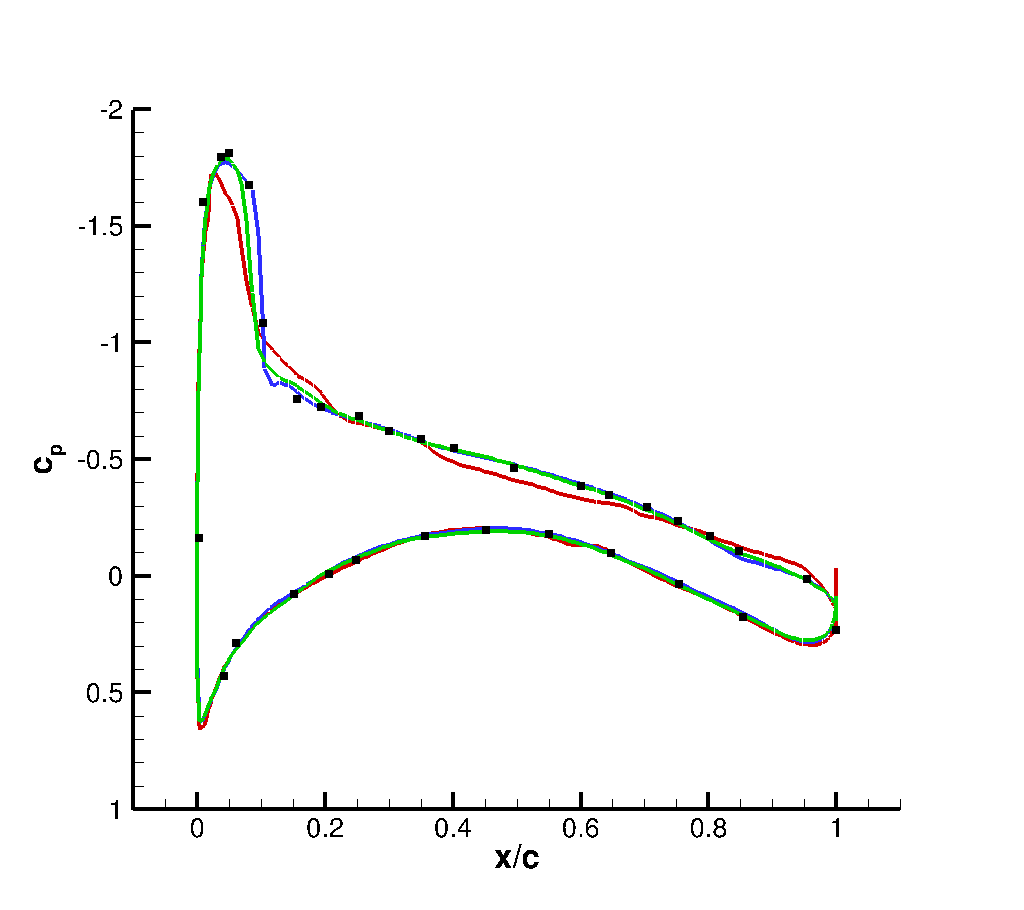}
    \caption{$\alpha = 4.0\si{\degree}$, section 3.}
    \label{fig:first}
\end{subfigure}
\hfill
\begin{subfigure}{0.49\textwidth}
    \includegraphics[width=\textwidth]{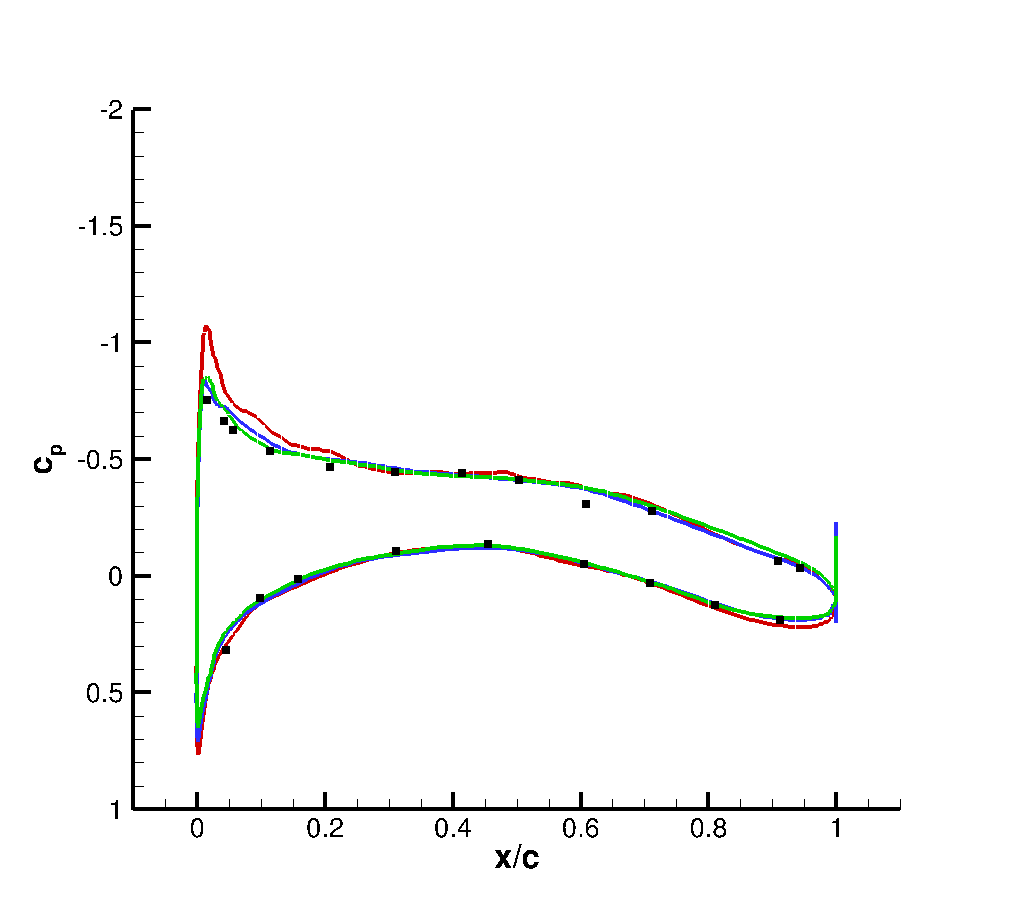}
    \caption{$\alpha = 4.0\si{\degree}$, section 9.}
    \label{fig:second}
\end{subfigure}
\caption{Pressure distribution in sections 3 and 9 for $M_\infty=0.70$ and $\alpha = \{2.0, 4.0\}\si{\degree}$.}
\label{fig:validation_exp_section_cuts_ma070}
\end{figure}

The pressure distributions for $M_\infty=0.7$ are shown in Fig.~\ref{fig:validation_exp_section_cuts_ma070}. The upper row shows the results for $\alpha=\SI{2}{\degree}$ in sections 3 and 9. At the innermost section, all the methods agree well with the measurement data. However, the gappy POD deviates from the measurements in section 9 where it overpredicts the pressure at the leading edge. In contrast, the neural network accurately describes the measurements in this region, showing that it is capable of extrapolating in the spanwise direction. The network trained by the multi-point strategy can also extrapolate in parameter space, as can be seen for $\alpha=\SI{4}{\degree}$ in the lower part of Fig.~\ref{fig:validation_exp_section_cuts_ma070}. The pressure distribution in section 3 features a distinct shock close to the leading edge which is accurately captured with both training strategies. As previously observed, the gappy POD has difficulties reproducing the shock leading to significant discrepancies on the whole suction side. Section 9 does not exhibit a shock but the gappy POD again overestimates the pressure at the leading edge.

\begin{figure}[b!]
\centering
\centering
\begin{subfigure}{0.49\textwidth}
    \includegraphics[width=\textwidth]{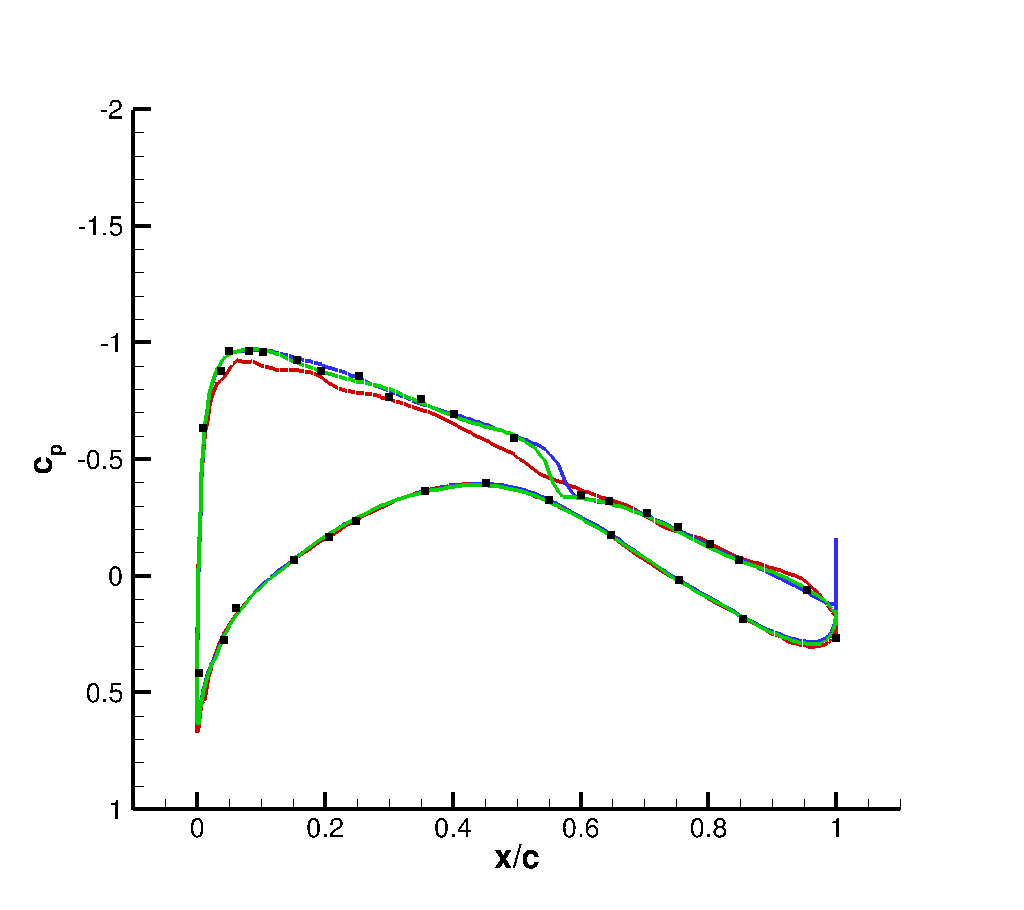}
    \caption{$\alpha = 2.0\si{\degree}$, section 3.}
    \label{fig:first}
\end{subfigure}
\hfill
\begin{subfigure}{0.49\textwidth}
    \includegraphics[width=\textwidth]{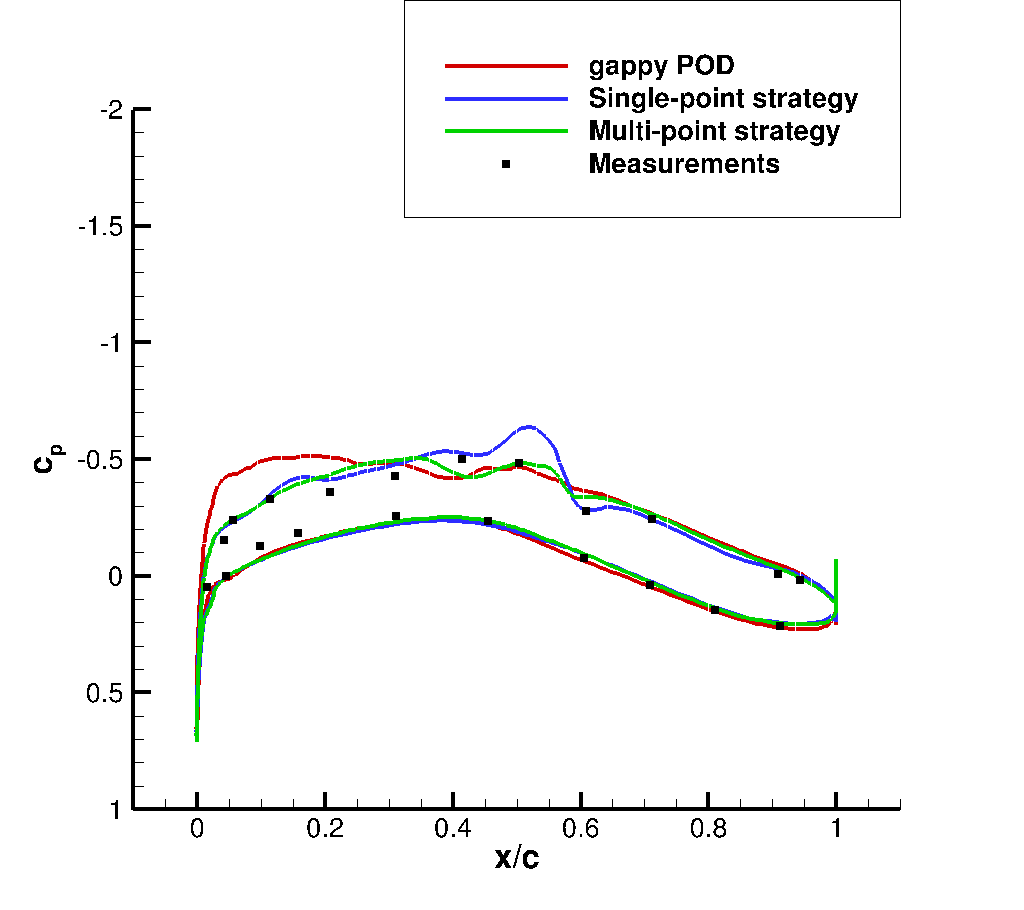}
    \caption{$\alpha = 2.0\si{\degree}$, section 9.}
    \label{fig:second}
\end{subfigure}
\begin{subfigure}{0.49\textwidth}
    \includegraphics[width=\textwidth]{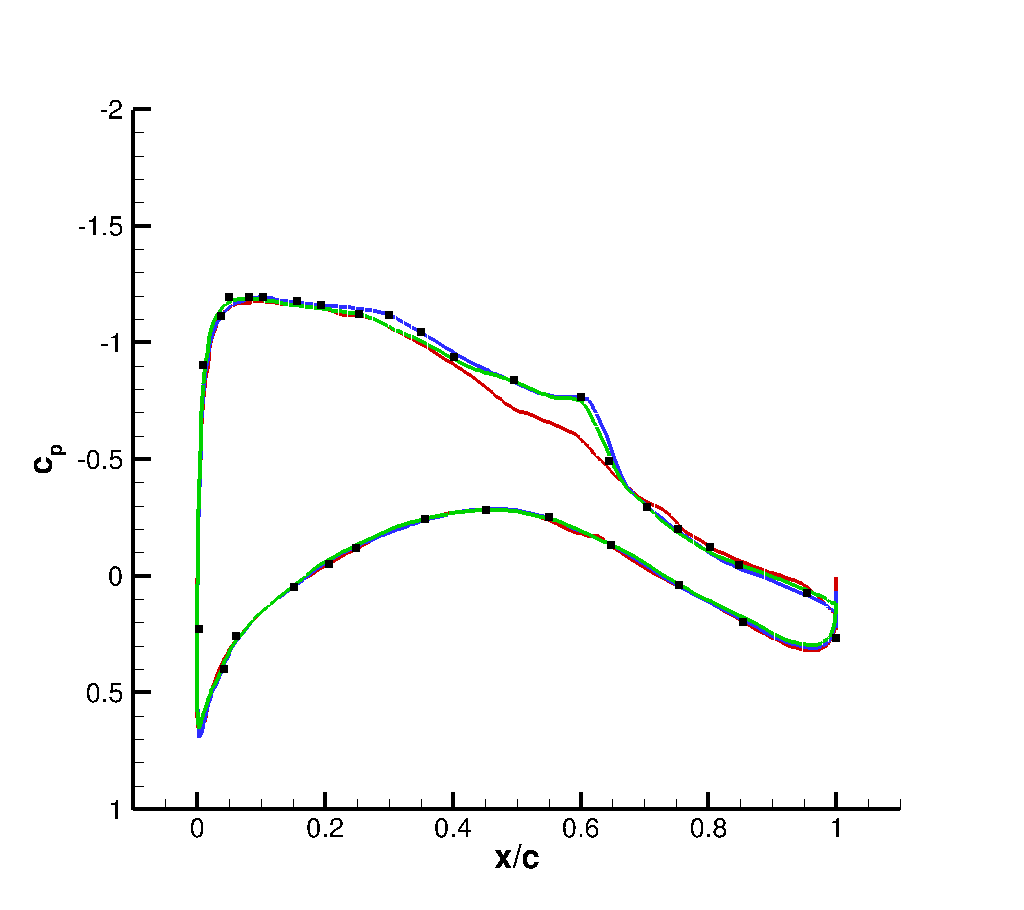}
    \caption{$\alpha = 4.0\si{\degree}$, section 3.}
    \label{fig:first}
\end{subfigure}
\hfill
\begin{subfigure}{0.49\textwidth}
    \includegraphics[width=\textwidth]{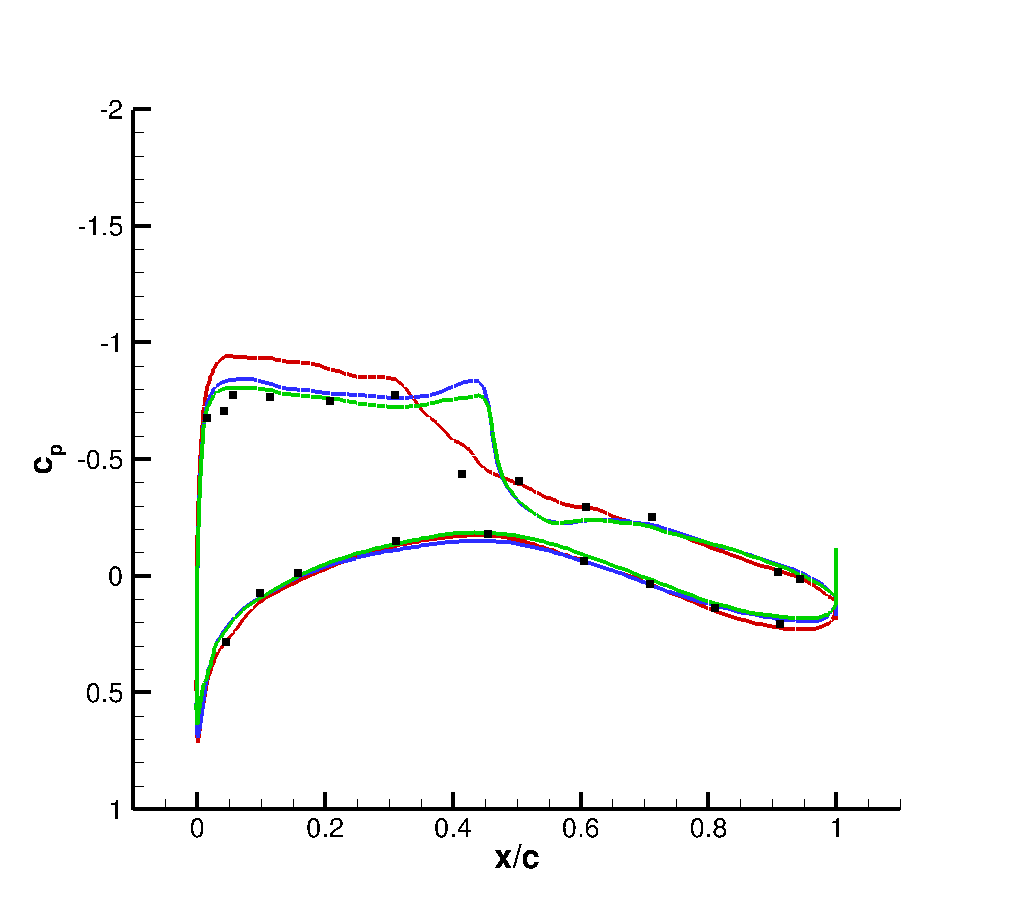}
    \caption{$\alpha = 4.0\si{\degree}$, section 9.}
    \label{fig:second}
\end{subfigure}
\caption{Pressure distribution in sections 3 and 9 for $M_\infty=0.85$ and $\alpha = \{2.0, 4.0\}\si{\degree}$.}
\label{fig:validation_exp_section_cuts_ma085}
\end{figure}

Figure~\ref{fig:validation_exp_section_cuts_ma085} depicts the pressure distributions for $M_\infty=0.85$, which is dominated by shocks. Hence, the gappy POD produces inaccurate results, especially in section 3 where the predictions show no signs of a shock. The predictions of the neural network retain the shock, resulting in very accurate results. The pressure distribution in section 9 is more challenging to model, as the methods have to extrapolate. As a consequence, the neural network predictions deviate in shock strength and position from the measurements. Both training strategies lead to similar results, illustrating a good generalization of training the multi-point strategy. Similarly to the results for $M_\infty=0.7$, the gappy POD overpredicts the pressure at the leading edge.


\section{Conclusion and Outlook}\label{sec:conclusion}

This work introduces a data fusion methodology for numeric and experimental aerodynamic data based on neural networks and transfer learning. The aim is to combine the high spatial resolution of the simulations with the high accuracy of the measurements. As both data sources are usually not available at the same spatial locations, the chosen network uses a mesh-free formulation where the network inputs are the spatial coordinates and the flow conditions. The training of the network is performed in two steps. In a first step, the network is pre-trained on an extensive numerical data set to learn relevant spatial features of the flowfield. In the second step, the model is fine-tuned on the measurement data to compensate for discrepancies. We compare two training strategies for the fine-tuning step. For the single-point strategy, fine-tuning and prediction is performed for the same flow condition and for the multi-point strategy, the model is fine tuned on multiple flow conditions and predicts for an unseen flow condition. The multi-point strategy has a clear advantage over the established gappy POD approach, which can only predict for the same flow conditions.

In the first step, the method is applied to synthetic measurement data extracted from high-fidelity simulation data of the NASA common research model. This allows us to quantify the accuracy of the fused predictions on the whole surface. While the neural network has an accuracy comparable to the gappy POD for subsonic cases, it performs better in transonic cases. In the presence of shocks, the gappy POD produces nonphysical oscillations, whereas the neural network accurately reproduces the shock by retaining the physical features of the simulations. Consequently, the neural network leads to smaller errors over all the tested flow conditions. Both training strategies result in similar error metrics, illustrating that the multi-point strategy generalizes well from a few measurement data points. In the second step, the method is applied to measurement data from the European Transonic Wind Tunnel. Two measurement sections are left out during training for testing. The neural network outperforms the gappy POD for subsonic and transonic cases, indicated by better extrapolation capabilities and more accurate shock capturing. The neural network shows good interpolation and extrapolation results using the multi-point strategy, indicating that it efficiently adapts the model to overcome the modeling errors of the simulations.

In summary, neural networks are a promising method for combining simulation and experimental data in a single model. As the transfer learning approach is very general, it can applied to more advanced neural network architectures, e.g. graph neural networks, potentially increasing the performance.
However, the chosen approach in this work assumes the measurements as ground truth, although they usually feature several sources of uncertainty. Hence, it would be favorable to include expert knowledge about this uncertainty during training. Additionally, uncertainty bounds for the prediction would be helpful in identifying regions where to trust the model and where not.

\section*{Acknowledgments}
The research leading to these results was partially funded by the German Federal Ministry for Economic affairs
and climate action (BMWK) as part of the LuFo VI-2 projects MuStHaF-DLR (funding reference: 20A2103C) and VIRENFREI ("Virtuelle Entwurfsumgebung für Reale, Effiziente Ingenieursleistungen", funding reference: 20X2106B).

The authors gratefully acknowledge the scientific support and HPC resources provided by the German Aerospace Center (DLR). The HPC system CARA is partially funded by 'Saxon State Ministry for Economic Affairs, Labor and Transport' and 'Federal Ministry for Economic Affairs and Climate Action'.

\bibliographystyle{elsarticle-num}
\bibliography{references}
\end{document}